%% file: main.tex
\documentclass[sigconf]{acmart}

\usepackage{booktabs} 
\usepackage{hyperref}
\usepackage{amsmath}
\usepackage{enumitem}
\usepackage{todonotes}
\usepackage{soul}
\usepackage[section]{placeins}

\usepackage[subrefformat=parens,labelformat=parens]{subfig} 
\usepackage{confcopyright} 


\setcopyright{rightsretained}

\acmDOI{10.475/123_4}

\acmISBN{123-4567-24-567/08/06}

\acmConference[SIGSPATIAL'18]{ACM SIGSPATIAL 2018}{November 2018}{Seattle, Washington, USA}
\acmYear{2018}
\copyrightyear{2018}

\acmArticle{4}
\acmPrice{15.00}

\begin{document}
\title{Trajectory annotation using sequences of spatial perception}

\author{Sebastian Feld}
\affiliation{\institution{Mobile and Distributed Systems Group\\LMU Munich}}
\email{sebastian.feld@ifi.lmu.de}

\author{Steffen Illium}
\affiliation{\institution{Mobile and Distributed Systems Group\\LMU Munich}}
\email{steffen.illium@ifi.lmu.de}

\author{Andreas Sedlmeier}
\affiliation{\institution{Mobile and Distributed Systems Group\\LMU Munich}}
\email{andreas.sedlmeier@ifi.lmu.de}

\author{Lenz Belzner}
\affiliation{\institution{MaibornWolff\\Munich}}
\email{lenz.belzner@maibornwolff.de}

\renewcommand{\shortauthors}{S. Feld et al.}

\begin{abstract}
In the near future, more and more machines will perform tasks in the vicinity of human spaces or support them directly in their spatially bound activities. In order to simplify the verbal communication and the interaction between robotic units and/or humans, reliable and robust systems w.r.t. noise and processing results are needed. This work builds a foundation to address this task. By using a continuous representation of spatial perception in interiors learned from trajectory data, our approach clusters movement in dependency to its spatial context. We propose an unsupervised learning approach based on a neural autoencoding that learns semantically meaningful continuous encodings of spatio-temporal trajectory data. This learned encoding can be used to form prototypical representations. We present promising results that clear the path for future applications.

\end{abstract}

\ACMCopyright{2018}{
    "Proceedings of the 26th ACM SIGSPATIAL International Conference on Advances in Geographic Information Systems",
    Association for Computing Machinery,
    New York,
    NY,
    USA,
    p. 329-338
}{10.1145/3274895.3274968} 

%
%



\copyrightyear{2018}
\acmYear{2018}
\setcopyright{acmcopyright}
\acmConference[SIGSPATIAL '18]{26th ACM SIGSPATIAL International Conference on Advances in Geographic Information Systems}{November 6--9, 2018}{Seattle, WA, USA}
\acmBooktitle{26th ACM SIGSPATIAL International Conference on Advances in Geographic Information Systems (SIGSPATIAL '18), November 6--9, 2018, Seattle, WA, USA}
\acmPrice{15.00}
\acmDOI{10.1145/3274895.3274968}
\acmISBN{978-1-4503-5889-7/18/11}

\maketitle

\input{body}

\bibliographystyle{ACM-Reference-Format}
\bibliography{thesiscites}

\end{document}

%% file: body.tex
\input{introduction}
\input{related_work}

\input{basics}
\input{concept}

\input{results}
\input{conclusion}

%% file: introduction.tex
\section{Introduction}
\label{sec:introduction}


Mobile robots enter our daily lives, be it in private or business context, to raise either productivity or comfort. Probably the most popular use case for such autonomous acting hard- and software systems is way-finding support in complex public environments like airports \cite{ruppel2009indoor}, fairs, or hospitals \cite{cosma2004autonomous}. Even daily-routine support in private environments like high-income households or home for the elderly (i.e., Active Assisted Living (AAL) \cite{rashidi2013survey}) may benefit from intelligent and autonomous mobile robots. Another important use case is Simultaneous Location and Mapping (SLAM), in particular in unknown or hazardous environments \cite{nagatani2013emergency}.

Originating from cognition psychology, spatial perception describes attempts to understand the environment human beings or other entities in general are surrounded by \cite{de2007geospatial}. Space syntax \cite{hillier1989social} are corresponding techniques to measure and analyze such local environments with isovist analysis \cite{tandy1967isovist, benedikt1979take} as a popular implementation focusing on all points visible from a given point of view. 
Besides analyzing a single point in space, one can also measure the spatial perception along a trajectory. The idea is that while continuously moving through space, one may also measure continuous changes in corresponding spatial perception.
There are several attempts utilizing space syntax techniques that define and analyze (psychology, e.g. \cite{wiener2004isovists}) or recognize and learn (computer science, e.g. \cite{sedlmeier2018discovering}) recurrent although fuzzy structures inside buildings like rooms, halls, or combinations of them.
Thus, transforming visual sensor input into some kind of spatial-temporal awareness may help creating human-machine-interfaces and wayfinding-systems with special requirements, e.g. the identification and presentation of routes for visually impaired persons that avoid identified hazardous situations \cite{li2004architecture, hub2004design}. Even SLAM units may benefit from spatio-temporal awareness when enabled to communicate their situational understanding or identified patterns in spatial perception along their path.
The machine learning community made huge advances creating techniques that may be suitable in the contexts mentioned above. There are numerous results in information extraction and pattern recognition on large scale and unknown data sets. Regarding the visual impression of space, Convolutional Neural Networks (CNNs) are able to detect patterns in images, i.e. visual imagery. Concerning the aspect of time, Recurrent Neural Networks (e.g. LSTM or GRU) or temporal convolutions are able to detect temporal dependencies in given data. Finally, the latent representation learned by some generative models, such as Variational Auto-Encoders (VAE), can be used to create low-dimensional informative representations of factors of variation in the data distribution of interest, such as trajectory data.


We hypothesize that it is possible to cluster movement through buildings based on spatial perception. Thus, the question to be answered is how to implement, train, and apply an artificial neural network in an unsupervised way using isovist sequences along trajectories through spatial structures in 2D worlds as input. More precisely, we suppose that a neural network is able to learn recurring spatio-temporal patterns in sequences of bitmaps of isovists and to cluster them for further usage (e.g., annotation).
As mentioned above, such a framework may evaluate and interpret environmental information and communicate it in a human-like way. The identification of patterns in an entity's spatial perception along a path may support the development of human-machine-interfaces by, for example, the annotation of trajectories based on the identified spatial perception.


After discussing related work regarding semantic annotation floor plans in Section \ref{sec:related_work}, we describe our methods and background in Section \ref{sec:background}. We then propose the concept of our system that is able to analyze movement through space in Section \ref{sec:algo}. Basically, it consists of (a) data synthesis, thus the creation of isovists along trajectories through floor plans, (b) the neural network architecture including CNN, GRU, and VAE, and finally (c) some visualization aspects. Section \ref{sec:ResultsEvaluation} incorporates an in-depth analysis of our system including the evaluation and discussion of several results. We conclude our paper in Section \ref{sec:conclusion} and briefly give hints on future work.

%% file: related_work.tex
\section{Related Work}
\label{sec:related_work}






This section contains related work regarding the semantic annotation of floor plans. Basically, floor plans constitute a subset of map representations of spatial environments, an important concept inside the communities of GIS and LBS. There are many different types of map representations, ranging from geometrical or topological to logical maps.

There is a huge corpus of existing literature regarding the semantic annotation of floor plans in the context of SLAM \cite{leonard1991simultaneous}. There are techniques that detect rooms and doors in order to create topological maps using virtual sensors, 2D laser scans, or camera images \cite{buschka2002virtual, anguelov2004detecting, chen2014door}. Further work semantically annotate maps using supervised learning techniques creating labels like room, corridor, hallway, doorway, or free-space \cite{mozos2006supervised, mozos2010semantic,goerke2009building}.
We delimit our work regarding three facts: (1) we focus on the analysis of floor plans without the integration of further sensors, (2) we focus on the impression of movement through space and not on the analysis of space itself, and (3) we follow an unsupervised and fuzzy approach.

Besides, there is related work that incorporates isovists for the analysis of architectural space, just as the paper at hand.
\cite{bhatia2012analyzing} estimate salient regions, i.e. regions with strong visual characteristics, in architectural and urban environments using 3D isovist.
\cite{feld2016approximated} approximate isovist measures along trajectories in order to identify traversed doors, for example.
Finally, \cite{feld2017identifying} calculate and cluster isovist measures on 2D floor plans showing that the identified clusters correspond to regions like e.g., streets, rooms, or hallways.
These approaches, however, do not take the impression of movement into account.

The most significant related work may be \cite{sedlmeier2018discovering}, where a framework for creating 2D isovist measures along trajectories traversing a 3D simulation environment is presented. The authors show that these isovist measures reflect the recurring structures found in buildings and that the recurring patterns are encoded in a way that unsupervised machine learning is able to identify meaningful structures like rooms, hallways and doorways. The labeled data sets are further used for neural network based supervised learning. The models generated this way do generalize and are able to identify structures in different environments.
Again, our paper delimits regarding the fact that we focus on the analysis of spatial perception during movement.

%% file: basics.tex
\section{Methods \& Background}
\label{sec:background}

The main goal of this paper is the clustering of interior movement using sequences of spatial perception. Thus, this section will describe methods and background used in this concept. We describe isovists as a representation of spatial perception in Section \ref{subsec:data_background}, followed by several machine learning techniques that focus on spatial and temporal pattern recognition as well as on unsupervised clustering (Section \ref{subsec:nna_background}).

\subsection{Environments and Isovists}
\label{subsec:data_background}

In GI-science, the definition of an environment ranges from a landscape-sized objects to space as a social construct and further. In general, it can be seen as being an immovable object with a surrounding character, while its surfaces give structure to an observer's topological perceived immersion.
We consider an environment as a three-dimensional finite structure that consists of a walkable floor, any kind of obstacles forming a boundary to the structure, and a ceiling as a closure to this construct. From an observer's perspective at any given point in this environment, the perceived visual space would be the area which can be described by all directly visible surfaces.

\begin{figure} 
	\centering
	\subfloat[Continuous Isovist]{
		\includegraphics[width=.42\linewidth]{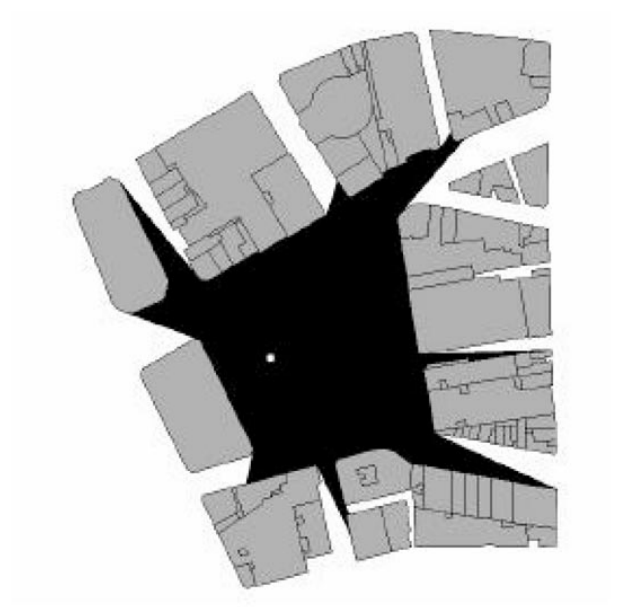}
		\label{fig:continious_isovist}
	}
	\subfloat[Discrete Isovist]{
		\includegraphics[width=.42\linewidth]{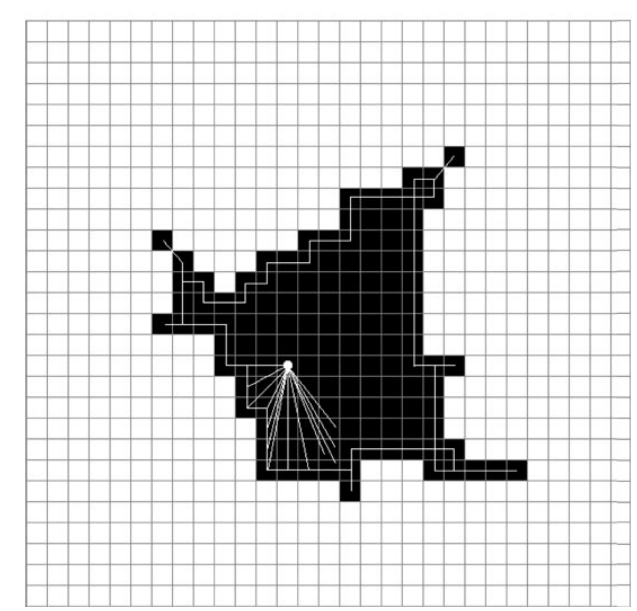}
		\label{fig:discrete_isovist}
	}
	\caption{\textit{Batty} \cite{batty2001exploring} demonstrates the difference between a continuous vector-based isovist model to a discrete grid-based one.}
	\label{fig:continious_discrete_isovist}
\end{figure}

Isovist analysis is a method first introduced by \textit{Tandy} \cite{tandy1967isovist} in 1967, which was later extended by \textit{Benedikt} \cite{benedikt1979take} in 1979. It transforms the perception of space into a measurable representation (cf. \autoref{fig:continious_discrete_isovist}). Isovist analysis is focused on retrieving and analyzing quantitative environmental information (structures and arrangements), rather than qualitative object attributes (texture, color, movability, function). 
\textit{Benedikt}\cite{benedikt1979take} describes the \textit{isovist} as ``location-specific patterns of visibility'' \cite[p. 7]{benedikt1979take}.
Thus, each \textit{isovist} describes the spatial perception at a specific position whereas a chronology of \textit{isovists} describes the spatial perception during a movement along multiple points in space. Since \textit{isovists} make spatial descriptions measurable and comparable, a sequence of \textit{isovists} makes a series of spatial descriptions during motion through space comparable. In literature, the concept of tracking motion using isovist analysis is referred to as \textit{isovist fields} \cite{batty2001exploring}. 

Besides isovist analysis there exist other methods to measure and process visible space. In \cite{llobera2003extending}, \textit{Llobera} gives an overview of available methods in visual analysis. In geographical and archaeological context, the concept of \textit{viewshed} captures visual space of the scale of natural landscapes by \textit{Digital Elevation Models} (\textit{DEM}). Furthermore, the concept of \textit{Visual Graphs} arises from the definition of ``isovists as a subgraph of a visibility graph''. It allows the calculation of measurable properties such as \textit{distance, area, perimeter, compactness, and cluster ratio} to be calculated and mapped back into space \cite{llobera2003extending, turner2001isovists}. \textit{Llobera} defines \textit{visualscapes} as an extension to Benedikt's initial ideas \cite{llobera2003extending}.

\subsection{Machine Learning \& Neural Networks}
\label{subsec:nna_background}

Our system consists of machine learning techniques that are able to describe visited training samples and build a classifier.

\subsubsection{Spatial Patterns \& Convolutional Neural Networks}
\label{subsec:cnn_background}

A CNN tries to learn and reveal spatial patterns by applying several filters and local pooling to an input. It basically consist of three layer types that are combined \cite{o2015introduction}. First, the convolutional layer can be described as a local connected weight multiplier. Small areas of its input are multiplied with an internally stored weight matrix on multiple filter levels. Each of those filters becomes specialized on a certain characteristic, like a discrete color or a spatial pattern. A feature is found through a high magnitude outcome of the filter's weight multiplication operation in relation to neighboring pixels \cite{lacey2016deep}. In detail, the convolution operation itself is achieved by calculating the scalar product within a kernel window (e.g., 3x3), which is moved over the input matrix rows and columns.

Additionally, pooling layers act as dimensionality reducers and merge semantic similar features into one feature using an arithmetic function. The most common kind of pooling is \textit{max pooling}, which works by splitting the input in (usually non-overlapping) patches and outputting the maximum value of each patch \cite{lecun2015deep}. As a provider of invariance, pooling operations also reduce a model's computational complexity \cite{o2015introduction, dumoulin2016guide}.

Lastly, a fully-connected layer usually serves as the final stage of a CNN. By connecting every neuron of the previous layer with each of the current layer's neurons, it attempts to produce class scores which can be used for classification (usually by applying the softmax function). Its way of operation is determined by the chosen activation function and structural position within an neural network \cite{o2015introduction}.

Since isovists can be processed in form of spatially correlated binary images, CNNs are an adequate tool for finding recurrent structures and natural patterns.

\subsubsection{Temporal Patterns \& Gated Recurrent Units}
\label{subsec:gru_background}

We assume a temporal relation within a sequence of isovist images. Sequential data input can be processed using \textit{Recurrent Neural Networks} (RNN).
While one element of a sequence is processed at a time, a hidden vector carries the history of all past elements of a sequence so that the output at a time step is the result of each previously evaluated input combined with the current input.
Thus, RNNs are comparable with a single layer that is reused multiple times in one iteration while tracking all subsequent computations \cite{lecun2015deep}.

While regular RNNs are known for their problems in processing long time dependencies (the learning gradient is often known to either explode or vanish \cite{lecun2015deep,chung2014empirical}) \textit{Hochreiter \& Schmidhuber} introduced the \textit{Long-Short-Term-Memory (LSTM)} unit which performes better on distant temporal relations \cite{hochreiter1997lstm}. The main difference between regular RNNs and LSTM units lies in the unit's connection to itself at the next time step through a memory cell and the introduction of a forget gate \cite{lecun2015deep}. Recently, \textit{Cho et al.} proposed \textit{Gated Recurrent Units} (GRU) as a modified LSTM unit that delivers comparable results at a lower number of weights/parameters.

Those units include a reset gate and an update gate that control how much each hidden unit remembers or forgets while reading a sequence \cite{cho2014gru}.

\subsubsection{Clustering \& Variational Auto-Encoder}
\label{subsec:ae_background}

Since human perception is highly subjective we need an unsupervised clustering approach that proves the assumption of a temporal and spatial relation within the training data in general.

\begin{figure*}
	\centering
	\includegraphics[width=\textwidth]{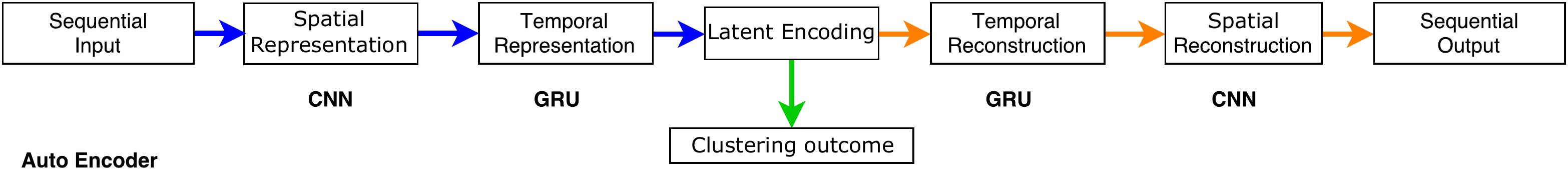}
	\caption{
		Overview of the NN architecture; \textbf{blue:} processing a sequential input through several layer to form an internal representation; \textbf{green:} network prediction; \textbf{orange:} building a sequential output as training target.}
	\label{fig:ml_overview}
\end{figure*}

The Auto-Encoder (AE) can be seen as a type of high-dimensional clustering network capable of learning prototypical representations. By comparing the previous network input to the generated output, an estimated error is calculated and afterwards back-propagated as the training signal (cf. \autoref{fig:ml_overview}) \cite{kamyshanska2015potential}. Its general concept consist of two stages: First, the probabilistic \textit{encoder} model \(q_\Phi(z|x)\) reduces an input's dimensionality and learns a representation. This representation vector is established in a small fully connected bottleneck layer (\autoref{fig:ml_overview}, blue). Second, the probabilistic \textit{decoder} or \textit{generator} model \(p_\theta(z|x)\) reconstructs the original input based on the given representation (\autoref{fig:ml_overview}, orange). During training, a loss function evaluates the decoder's output by comparing it to the encoder's input. Since AEs operate without the need for labeled data, the entire process is considered to be unsupervised.

Typically, the AE's layers consist of fully connected neurons, however, there are cases in which AEs are combined with other ANN structures like CNNs (Section \ref{subsec:cnn_background}) or recurrent-layers (Section \ref{subsec:gru_background}). It can be simply described as a deep discriminative ANN whose output targets are the data input itself rather than class labels \cite{deng2014deep, lecun2015deep}. In 2013, \textit{Kingma} \cite{kingma2013auto} introduced a learning enhancement called the Auto-Encoding Variational-Bayes algorithm, which allows for a better approximation of ``posterior interference using simple ancestral sampling'' \cite[1]{kingma2013auto}. \textit{Kingma} proposed the \textit{Variational Auto-Encoder} (\textit{VAE}) whose main benefit lies in the structure that can be discovered within the bottleneck layer. The resulting probability distribution of sample representations in latent space can be approximated to any desired choice. For instance, a Gaussian distribution for real-valued data or a Bernoulli distribution for binary data input is applicable.

As Auto-Encoders have shown to provide state-of-the-art performance in a variety of tasks like object recognition or learning invariance in representations \cite{kamyshanska2015potential}, we employ a VAE as the overall neural network structure capable of handling sequences of isovists.

%% file: concept.tex
\section{Concept}
\label{sec:algo}

Our system's main goal is the clustering of movement based on the visual impression during movement while following a path that runs through an interior space (e.g., a building). We therefore have presented our environment (Section \ref{subsec:data_background}) which is modeled as a discrete occupancy grid together with trajectories traversing it. For each visited position on a discrete trajectory we compute isovists and use sequences of such as our data basis, called an isovist-sequence. Such can therefore be used to analyze changes in spatial perception.

In Section \ref{subsec:nna_background} we have explained three key methods of machine learning used in this paper. We now propose the combination of Convolutional Neural Networks (CNN) and Gated Recurrent Units (GRU) elements embedded in an Auto-Encoder (AE) structure allowing to cluster temporal correlation within sequences of two-dimensional bitmaps as representations of spatial structures. Subsequently, we present the clustering outcome as a visualization of annotated trajectories in Section \ref{sec:ResultsEvaluation}.

\subsection{Data Synthesis}
\label{subsec:data}

In this work, machine perception is considered as an entity's capability to recognize spatio-temporal patterns based on a sequence of spatial sensory input in form of isovists, a technique that makes the description and perception of architectural space quantifiable and, thus, appropriate for utilization \cite{benedikt1979take, tandy1967isovist}. 
Based on the review of existing techniques in the field of visual space, we decided to use isovists as our measure of perception of space (Section \ref{subsec:data_background}).
In literature and application, the simplest and most often used form of \textit{isovists} is a 2D top-down floor plan \cite{llobera2003extending, batty2001exploring}. 
A problem reduction to 2D floor plans as source of spatial structures encoded in visible floor and non-visible floor is legitimate, since a robotic unit only needs information about the floor on which it can safely move.
Those binary images are composed of black pixels (non-visible floor, wall elements, or obstacles) and white pixels (visible and walkable floor).

Thus, our system's data preprocessing and input creation consists of three parts: (1) reading floor plans from binary images to build a routable graph, (2) generating paths to simulate motion/trajectories and (3) computing \textit{isovists} for each visited pixel position of a trajectory using a shadow casting algorithm \cite{bergstroem2017shadow, bergstroem2015FOV}. Just like \textit{Benedikt's} explanation of \textit{isovists}, the \textit{Shadow Casting}-algorithm spreads from a source pixel. This proceeding generates a sequence of binary isovists that are consecutively rotated in walking direction.

\begin{figure}
	\centering
	\includegraphics[width=.83\linewidth]{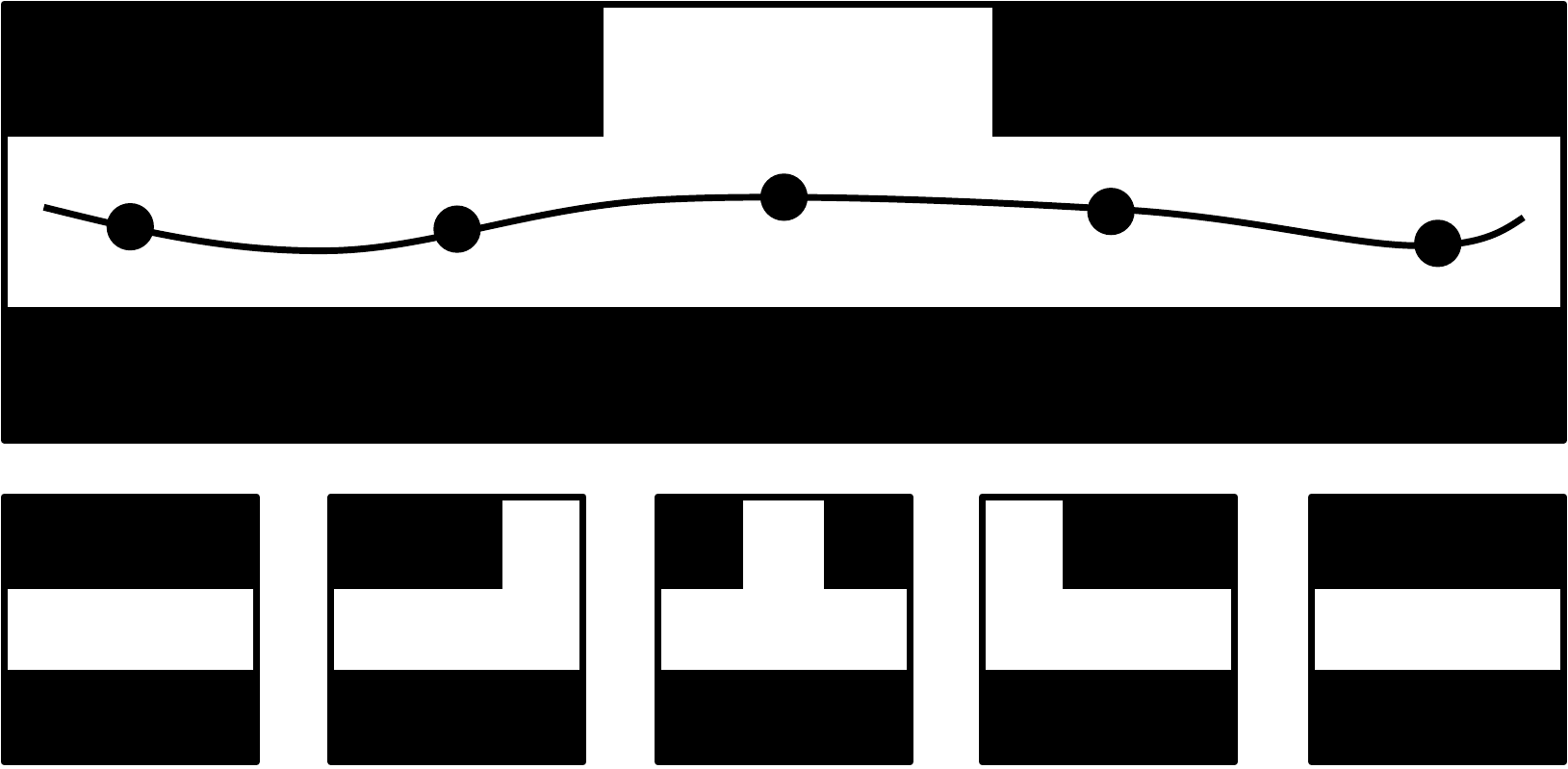}
	\caption{Schematic isovist-sequence setup.}
	\label{fig:is_setup}
\end{figure}

Depending on the use-case, instead of simulating the visual perception as described above,
it would also be possible to generate the isovists from real world sensory input, e.g. collecting measurements via laser scanning.
As there are various ways of collecting or generating the required isovists, the work at hand does not focus on the specifics (performance etc.) of the data synthesis or collection procedure.

Figure \ref{fig:is_setup} shows a schematic overview of our isovist-sequence setup.

\subsection{Neural Network Architecture}
\label{subsec:nna}

After introducing our understanding of environment and isovist-sequences as a tool to capture changes in spatial perception, we now propose an \textit{Artificial Neural Network} (ANN) architecture that clusters a sequence of temporal-related binary bitmaps of two-dimensional spatial representations.

We utilize several machine learning techniques to automatically discover a function by employing CNNs (Section \ref{subsec:cnn_background}) and GRU (Section \ref{subsec:gru_background}). Both are combined within an unsupervised trained VAE structure (Section \ref{subsec:ae_background}).

The first step of our architecture is the extraction of visual pattern. We chose to use a CNN structure as described in Section \ref{subsec:cnn_background} employing two convolution/pooling bundles with 10 filters at a $3x3$ kernel size at a stride of 1, activated by a $ReLU$ function. Since we assume a temporal relation within a sequence of isovist images, we use a single GRU layer equipped with 250 cells to process the sequential data input by a $tanH$ function. After detecting spatial and temporal patterns, the next logical step in our architecture's design aims at detecting classes or groups within the isovist sequences. As stated above, due to the high performance in a variety of clustering tasks, we eploy a VAE as the overal neural network structure. It has been implemented as described by \textit{Kingma et al.} using a variational constraint. Followed by a similarly built generative model (first GRU, then convolution/pooling bundles), we establish a reconstruction of the raw input data as a valid training target \(p_\theta(z|x)\). A cross entropy loss with applied Kullback-Leibner (KL) divergence to a Gaussian distribution is utilized to produce an error signal that can be back-propagated as the training signal to perform the network's internal weight adjustment \cite{kingma2013auto}.

After training, the VAE structure is not in need of the decoder model any more.
Thus, the ANNs bottleneck can be exposed as clustering result (\autoref{fig:ml_overview}, green).

Summarized, the ANN architecture has been implemented using Keras with a Tensorflow backend in Python \cite{tensorflow2015-whitepaper, python2017website, chollet2017kerasio}.

\subsection{Visualization}
\label{subsec:viz}

\begin{figure}
	\centering
	\includegraphics[width=.97\linewidth]{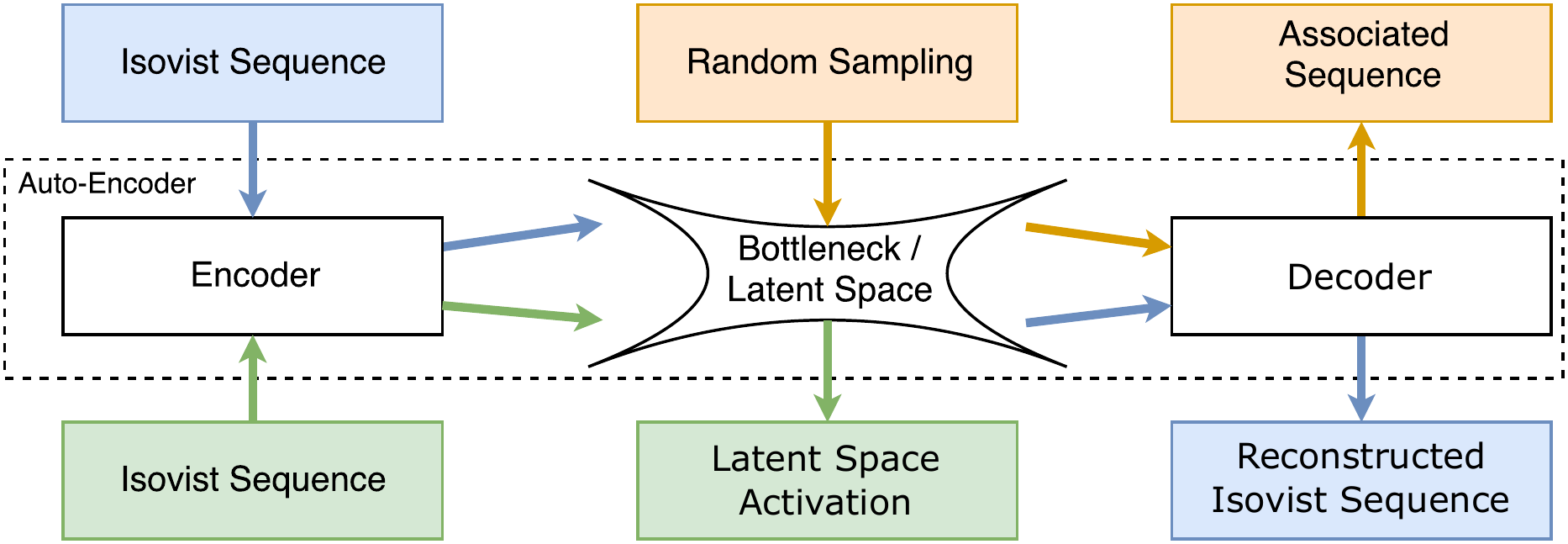}
	\caption{Visualization work-flow: comparing encoder input with the associated decoder output (\textbf{blue}), latent activation of encoder inputs (\textbf{green}), and isovist-sequence reconstruction (\textbf{orange}).}
	\label{fig:viz_workflow}
\end{figure}

In the previous sections we have introduced the overall structure of our network which can be trained in an unsupervised fashion. This section now introduces the visualization process that exploits the VAE's internal representation to annotate trajectories traversing a two-dimensional environment. Having an AE structure, we provide three possibilities of visualizing the network's outputs as pictured in \autoref{fig:viz_workflow}.

First, the encoder input (an isovist sequence) is viewed next to the related decoder output. This way, it is possible to validate the network's reconstruction capability visually to get insights into its way of function (\autoref{fig:viz_workflow}, blue). Second, by sampling from the compressed latent space in either random fashion or by using a steady pattern along a regular grid of the size of the latent dimension, associated sequences are revealed. In other words, the overall latent structure can be observed by feeding a synthetic latent vector to the generative model (orange). Third, isovist sequences along a trajectory are used for annotation by collecting latent vector activations. Those are afterwards drawn on a related floor plan to reveal patterns of spatial and temporal similarity (green).

%% file: results.tex
\section{Results \& Evaluation}
\label{sec:ResultsEvaluation}

We see our approach as a high-dimensional clustering concept using sequences of two-dimensional isovists in a general way to analyze changes in spatial perception. As such, it may support the implementation of sophisticated robotic navigation and assistance systems.

To proof our assumptions, we have implemented the proposed neural network architecture (Section \ref{sec:algo}) as a system that processes sequences of isovists. These isovists are in the form of a three-dimensional binary grid holding spatio-temporal relations. The network then produces fuzzy labels that can be used for trajectory annotation or other purposes.

\begin{figure}
\centering
\subfloat[]{
	\includegraphics[height=1.9cm,keepaspectratio]{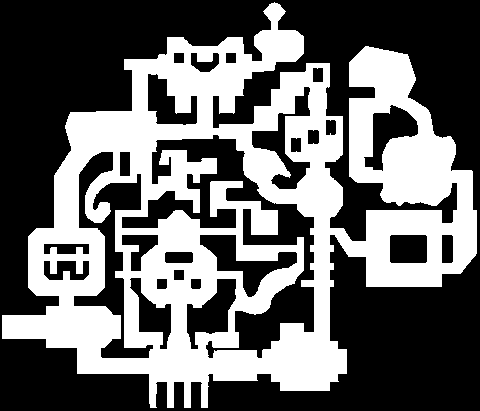}
	\label{fig:doom}}
\subfloat[]{
	\includegraphics[height=1.9cm,keepaspectratio]{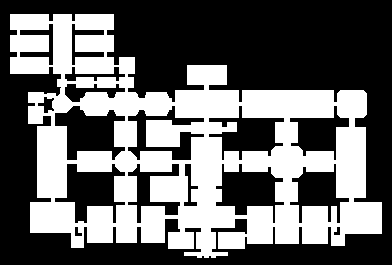}
	\label{fig:tate}}
\subfloat[]{
	\includegraphics[height=1.9cm,keepaspectratio]{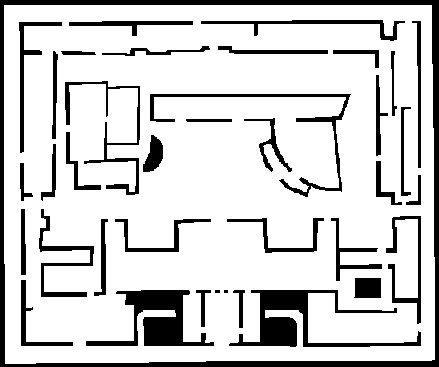}
	\label{fig:tum}}
\\ 
\subfloat[]{
	\includegraphics[height=1.5cm,keepaspectratio]{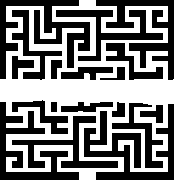}
	\label{fig:maze}}
\subfloat[]{
	\includegraphics[height=1.5cm,keepaspectratio]{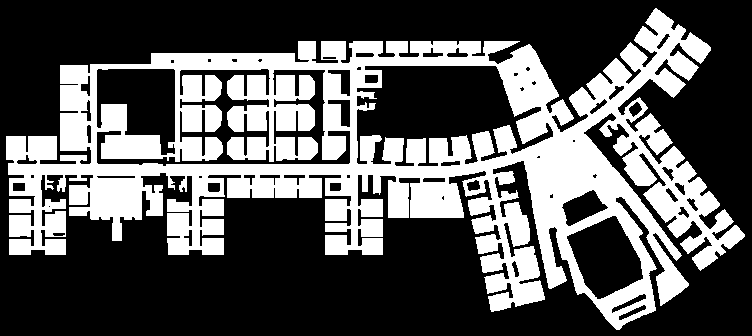}
	\label{fig:oet}}
\subfloat[]{
	\includegraphics[height=1.5cm,keepaspectratio]{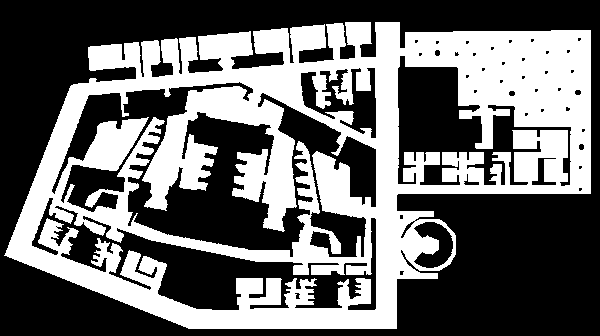}
	\label{fig:priz}}
\caption{Environments used in training.}
\label{fig:maps}
\end{figure}

We have evaluated the results of training the network on a total of six different floorplans (\autoref{fig:maps}) with varying shapes of rooms and hallways. This setup has been chosen to allow the network to learn generalizations across different floor plans and possibly generalize to unknown floorplans. Therefore, attention was given to include as many variations of room sizes, orientations, round \& square structures, for example, as possible. All floor plans were scaled to an equal average door width of four pixels. Besides, we build a pixel-wise routable graph and applied Dijkstra's algorithm \cite{dijkstra1959note} to generate several thousands of artificial paths through the environments. Our network has been trained on equally-sized isovist sequences having a length of $SequenceLength\ t=5$ that were separated by a $Spacing$ of $s=2$ pixels. The path segment covered by such a sequence is therefore $|IS| = t \times s - (s-1) = 9$.


There is no fixed number of necessary training repetitions, so called \textit{epochs}, that can be directly taken from literature. The correct amount depends on the number of samples, the variation along the data, and the problem's complexity. Experiments showed suitable results after about $100$ epochs visiting about $60~000~000~IS$ in training.

\begin{figure}
	\centering
	\includegraphics[width=.65\linewidth]{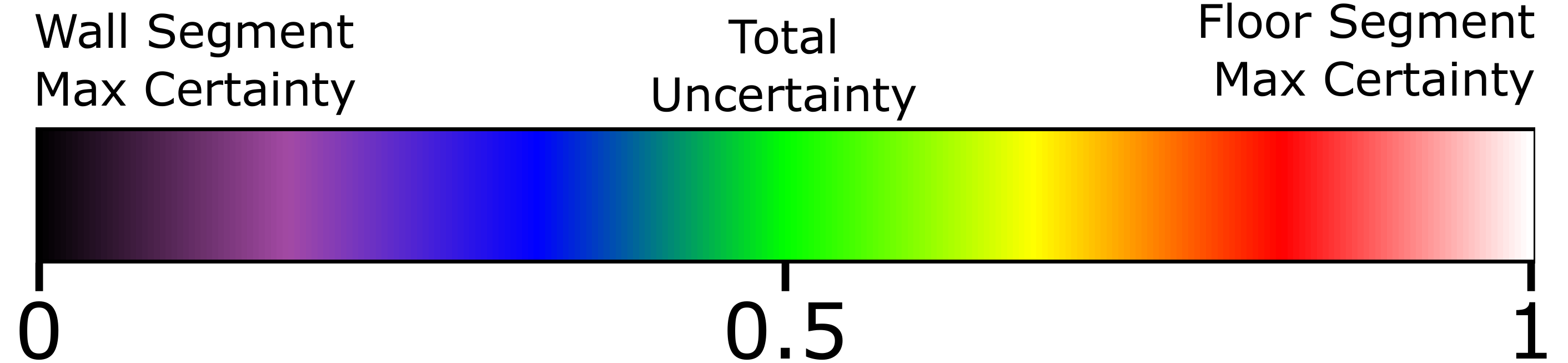}
	\caption{A custom colorbar represents the network's certainty on isovist-sequence reconstructions.}
	\label{fig:classcolors}
\end{figure}

Since this work deals with unsupervised clustering of perception without any labeled data, there is no recognized metric that can be applied ``out of the box''. We therefore performed a visual evaluation of results using the proposed methods of Section \ref{subsec:viz}. The desired outcome would be the analogous colorization of similar movement in a related spatial situation. For that, we have placed the network predictions at the center-pixel $P_{t}$ of an isovist-sequence from a longer artificial trajectory on an underlying floor plan (\autoref{fig:viz_workflow}, green). Additionally, VAE decoder predictions of evenly spaced latent samples are evaluated (\autoref{fig:viz_workflow}, orange). Those artificial \textit{IsovistSequences} can be used to understand a VAE's internal latent bottleneck structure.

For completeness, when choosing a higher latent space dimensionality, the output for tested samples could be subsequently clustered. Well established techniques like applying k-means clustering \cite{ball1965isodata} or the DBSCAN algorithm \cite{ester1996density} are only two of many possible post-processing options. For visual insights, a dimensionality reduction using \textit{Principle Component Analysis} (PCA) \cite{jackson2005user} or \textit{t-Distributed Stochastic Neighbor Embedding} (t-sne) \cite{maaten2008visualizing} could be applied on a set of latent space activations. 

The following results were generated by predicting on a new environment which was not part of the training data. Since it is hard to annotate multiple overlapping trajectories within a two-dimensional figure, several manually created paths are presented. Not only had the network to generalize from its learned spatial structure to an unknown environment, but also the process that formed the underlying trajectories was of foreign nature as well.

\subsection{Perceptive Trajectory Annotation}
\label{subsec:result_main}

In this section, we will demonstrate our system's annotation capabilities, the clustering outcome, and proof the correct encoding of temporal pattern along sequences of spatial representation captured by isovists. For that, we first show trajectories in a foreign environment that are annotated next to an evenly sampled latent space that results in sequential decoder reconstructions.

The color coding of such reconstructed isovist-sequences are shown in \autoref{fig:classcolors}. It runs from black (0, wall) over green (0.5, total uncertainty) to white (1, floor). A black colored pixel, for example, indicates a maximum certainty that this particular pixel had a black color when it entered the network, whereas a green pixel implies an even likelihood for both binary extremes. Such reconstructions can be visualized in order to gain insights into the learning process and to reveal the internal latent VAE structure. We picture them as horizontal bundles of mostly five isovists. The chronology is named from t$_{n-2}$ to t$_{n+2}$. Please mind the correct orientation along those visualizations. Sequence position indicators $t_n$ are always positions on the back side of an imagined agent's movement direction.

\begin{figure*}
	\centering
	\subfloat[A single cell VAE has been used to color hand drawn paths through the test environment. The training data parameters were set to \(t=5, s=2\).]{
		\includegraphics[width=.67\textwidth]{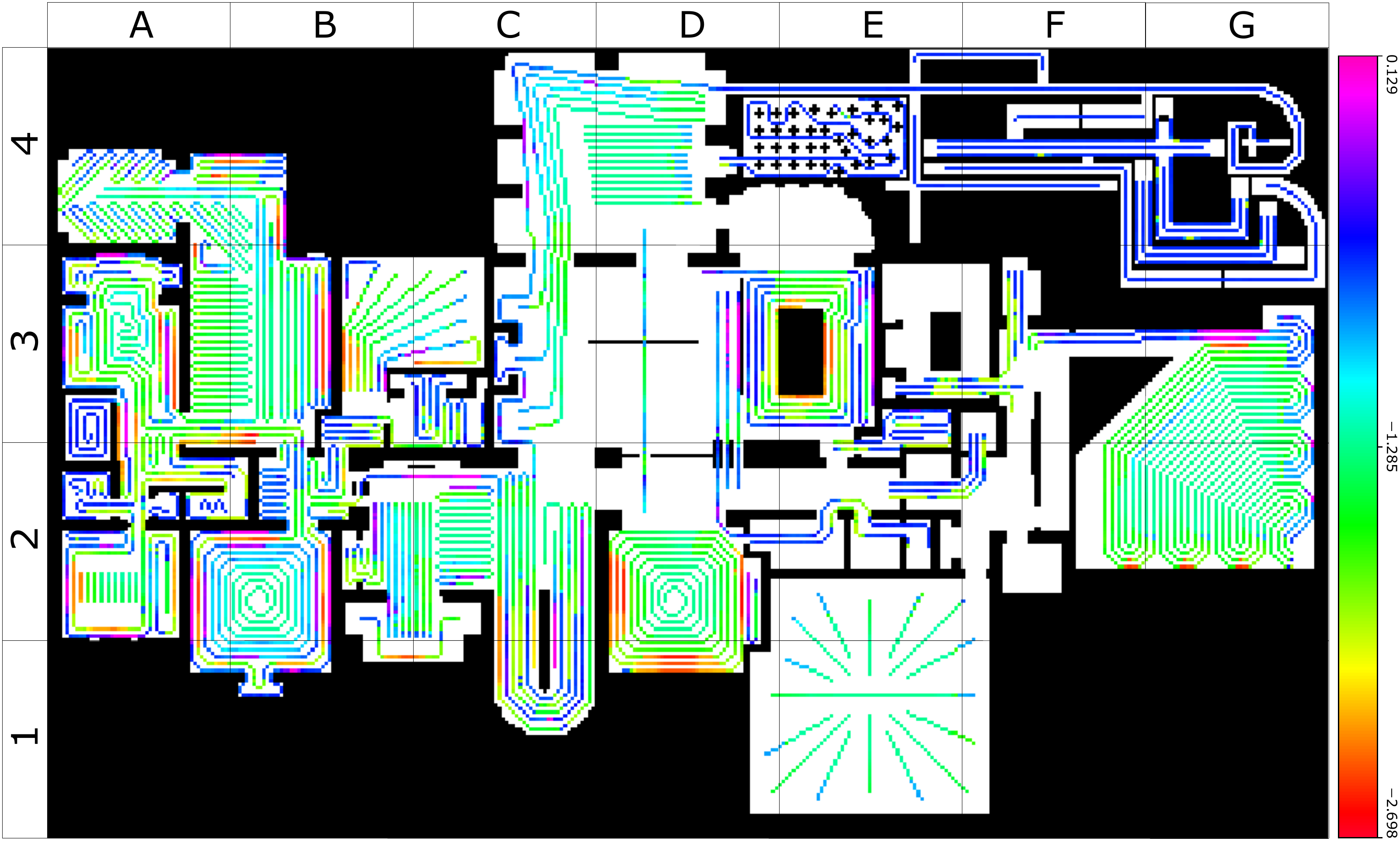}
		\label{fig:VAE_MAP_1_R}
	}
	\\ 
	\subfloat[Top: Equally spaced samples of the VAEs latent space used for the coloration of Figure \ref{fig:VAE_MAP_1_R}. Every column represents an isovist sequence from bottom $t=-2$ to top $t=2$. The walking direction points to the right; Bottom: Color map used to color hand drawn paths through the test environment.]{
		\includegraphics[width=.67\textwidth]{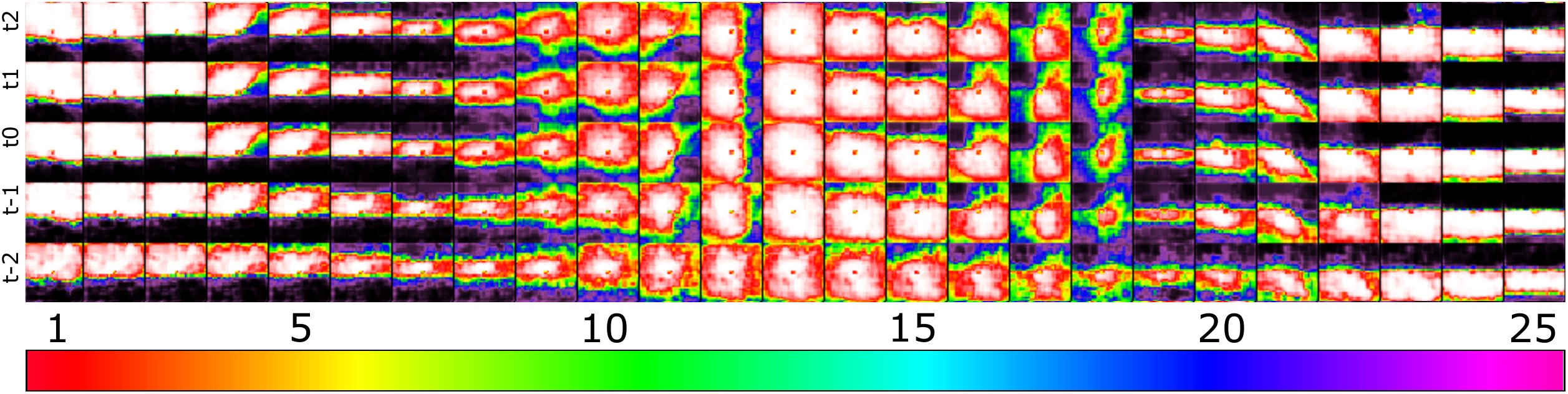}
		\label{fig:VAE_MAP_1_R_SPACE}
	}
	\caption{Pixel-wise annotation of trajectories based on spatio-temporal context.}
	\label{fig:main_results}
\end{figure*}

Figure \ref{fig:VAE_MAP_1_R} shows the two-dimensional trajectory annotation results based on high-dimensional spatio-temporal clustering through VAE. For that, our model was used to predict consecutive isovist-sequences along hand drawn trajectories which have not been visited in training. For better orientation during reading, floor-plans have been divided in $4\times7$ areas referred to as \textbf{Sn}, where \textbf{S} refers to the horizontal letter and \textbf{n} refers to the vertical digit.

In Figure \ref{fig:VAE_MAP_1_R_SPACE}, a one-dimensional VAE's latent space has been sampled in a regular interval based on the predicted values in Figure \ref{fig:VAE_MAP_1_R}. Both color-bars are directly connected, thus, a color found in the base-map (Figure \ref{fig:VAE_MAP_1_R}) is therefore further explained by the sequence next to the same color in Figure \ref{fig:VAE_MAP_1_R_SPACE}. The custom trajectories on the test map were then evaluated sequentially by the VAE encoder. Resulting predictions have been normalized and were transformed into RGBA value tuples.

We now examine the evenly spaced latent space samples in Figure \ref{fig:VAE_MAP_1_R_SPACE} from left to right. Columns 1-5 describe movements along a right-handed wall (e.g., spiral at area \textbf{D1} in Figure \ref{fig:VAE_MAP_1_R}). Besides, there is a disruption similar to an approaching wall visible at column 4. Then, uncertainties and thick wall segments are joining in from the right. The RGBA color shifts from a strong red to a greenish tone per yellow. In the following, green to turquoise color tones describe a spatial situation under a lot of changes, like right-handed curves (column 11) or towards a wall (12). At column 13 the representation of a movement through completely free space can be found (e.g., area \textbf{E1} in Figure \ref{fig:VAE_MAP_1_R}). The following reconstructed sequences, colored in blue tones, cluster movements with left-handed curves. The very narrow environments (area \textbf{F4} \& \textbf{G4}, blue color, column 19) can be seen just before the corridor widens (column 20). Diagonal wall segments (pixel neighboring each other in 45\textdegree{}) and right-hand side free-space is visible (spiral at areas \textbf{A2/B2} in Figure \ref{fig:VAE_MAP_1_R}). The very last sequences of the sampled latent range represent straight corridors of various width. The colors shift from blue to purple and finally a pink tone.

In short, our model has been successfully applied to visualize the clustering of movement in varying spatial context. The additional latent sampling visualization (Figure \ref{fig:VAE_MAP_1_R_SPACE}) helped to get insights into unsupervised colored trajectories (Figure \ref{fig:VAE_MAP_1_R}). The main findings are the identification of movements through free-space (turquoise), along right-hand (red) and left-hand (purple) wall alignment as well as through narrow corridors (blue).

However, the sometimes not directly interpretable results indicate the necessity for further optimization. With an increase of the latent space dimensionality the overall generalization is assumed to be less strict and results in more meaningful results when colored accordingly or mapped to a semantic label.

\subsection{Temporal Layer Validation}
\label{subsec:result_motion_capture}

\begin{figure}
	\centering
	\includegraphics[width=\linewidth]{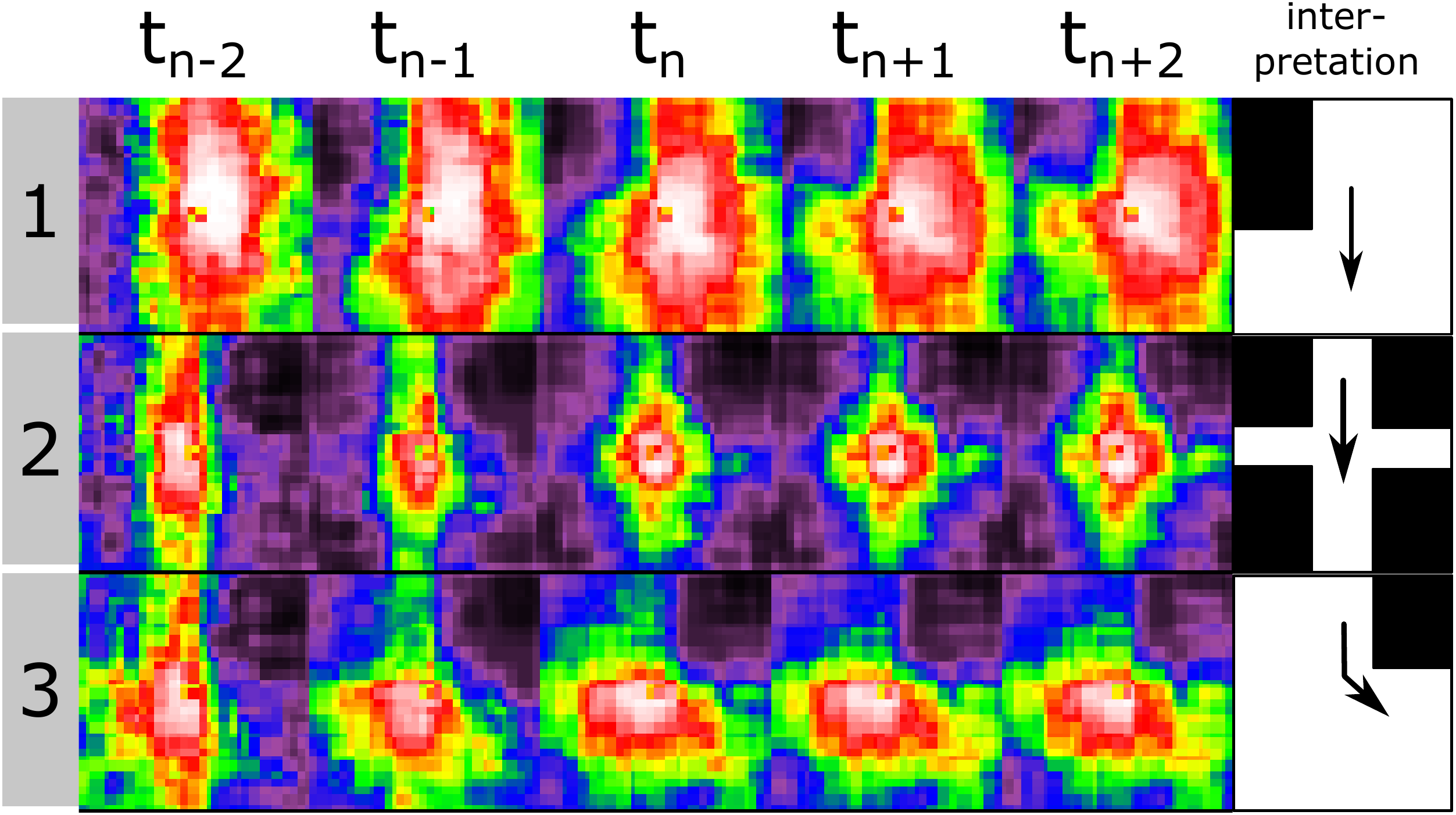}
	\caption{VAE samples showing temporal reconstruction along the time steps.}
	\label{fig:temporal_relation_conservation}
\end{figure}

We now proof our system's spatio-temporal clustering capabilities by observing decoder outputs such as the hand-picked examples in Figure \ref{fig:temporal_relation_conservation}.
Each of the rows represents a reconstructed isovist sequence with a clearly visible movement. Not only the spatial structure for each time step (t$_{n\pm2}$), but also the shifted center caused by an imagined virtual agent's movement in time $t$ has been successfully encoded. 
The second row of Figure \ref{fig:temporal_relation_conservation}, for example, can be read as a motion along a corridor approaching a crossing. 
This demonstration is essential, since it proves the correct application of the GRU units in combination with convolution and pooling layers.

More importantly it shows that temporal pattern are encoded, transferred through the small one-dimensional bottleneck-layer, and subsequently restored. Furthermore it can be concluded that these images demonstrate a successful learning and back-propagation behavior.

\subsection{Influence of Isovist Sequence Lengths}
\label{subsec:result_sequence_length}

In addition to the former results, we now present the influence of the \textit{sequence-length} parameter by increasing it to $t =9$. This also increased the length of our total observation to $|IS| = 17$. Thus, computational costs are almost doubled by introducing additional CNN and GRU operations to the network structure.
The results are pictured in Figure \ref{fig:VAE_MAP_1_T9}. Surprisingly, much less variations were differentiated in narrow situations (e.g., \textbf{A3} in Figure \ref{fig:VAE_MAP_1_T9}). On the other hand, areas including any kind of wide free-space are colored in more detail, as it can be observed in area \textbf{C2}, for example. Totally free-space is clearly represented by a purple color. Red and turquoise colors, on the other hand, seem to describe a movement along either a left- or right-handed wall in the context of free-space. The regular latent space sampling in Figure \ref{fig:VAE_1_T9_EachLatent} supports this assumption, as there are more white pixels visible than in direct comparison to Figure \ref{fig:VAE_MAP_1_R_SPACE}. 

The network generalized over smaller spatial features while, at the same time, differing much more along movement that involves some kind of free-space on either or both sides in the imagined agent's walking direction. 

\subsection{General Discussion}
\label{subsec:general_discussion}

This section discusses the lately presented results.

Our initial idea was the prediction of multiple isovist-sequences that were based on similar trajectories like those used in training. Thus, an additional trajectory-set filled with approximately $1\ 000$ trajectories had been constructed. The creation process was exactly the same as for the training data, i.e., we collected shortest Dijkstra paths from random start to random target coordinates. The emerging problem lays within the unavoidable overlap of such random shortest paths resulting in visually mixed trajectories.

Colorization of floor plans such as shown in Figure \ref{fig:maps} was first realized by the \textit{winner-takes-it-all} principle. Anyhow, this was only applicable in simple, straight corridors, when mostly similar movement types occurred. Crossings, or more general, spaces that allow various forms of movements, suffered from this method. Additionally, it is not possible to plot our system's prediction onto a single map pixel.

A trajectory can basically be seen as a positional change under the influence of time $Tr_{i=0}^\infty = [(x, y)_0,...,(x, y)_i ] : i \in \mathbb{N}$. A subset of a trajectory ($Tr$) follows the same definition $Tr_{t=m}^n = [(x,y)_m,...,(x,y)_{m+t}] : m,t \in \mathbb{N}$. Successively generated isovists, for such a $Tr$-subset of a specific length ($t$), results in our isovist-sequence. Thus, neural network predictions (vector of size n) represent spatio-temporal (three-dimensional) data inputs. 

Placing such a one-dimensional representation on an environmental pixel eliminates the context of the encoded movement in total. Only by marking the center positions of all involved isovist the movement context can be revealed again. Another solution is the prediction of multiple consecutive isovist-sequences that are part of a single trajectory. Using uneven sequence-lengths $t = 2m+1 : m\in \mathbb{N}$ while placing the prediction at the center positions $Tr_n$ where $n = |IS| \div 2 + 1$ showed the best results. 

As stated before, overlapping trajectories cannot be visualized in this way, which is why it was not possible to use the same random trajectory generation method as in training.
Instead, custom non-overlapping trajectories were generated by hand as a solution.
The selection of these custom trajectories aimed to fulfill two separate goals: ``squeezing'' as many non-overlapping paths on a floor plan as possible while at the same time having a high variation of spatial configurations, movement situations, and distances to the walls.
A drawback of this manual method is the possible introduction of human bias compared to the random Dijkstra's shortest path based trajectory generation procedure used in training.

As a result, not only the test-environment possesses totally new characteristics, the $Tracks$ do as well. As long as there is no applicable metric for measuring and comparing the performance on spatial perception clustering, this chosen method delivered results that are readable and allowed valuable insights. This course of action additionally tests the networks generalization capabilities. 

%% file: conclusion.tex
\section{Conclusion and Future Work}
\label{sec:conclusion}

\begin{figure*}
	\centering
	\subfloat[At parameter settings of \(t=9\) \& \(s=2\), the network generalized over smaller features while, at the same time, differing much more along sequences that involve some kind of free-space on either or both sides.]{
		\includegraphics[width=.67\textwidth]{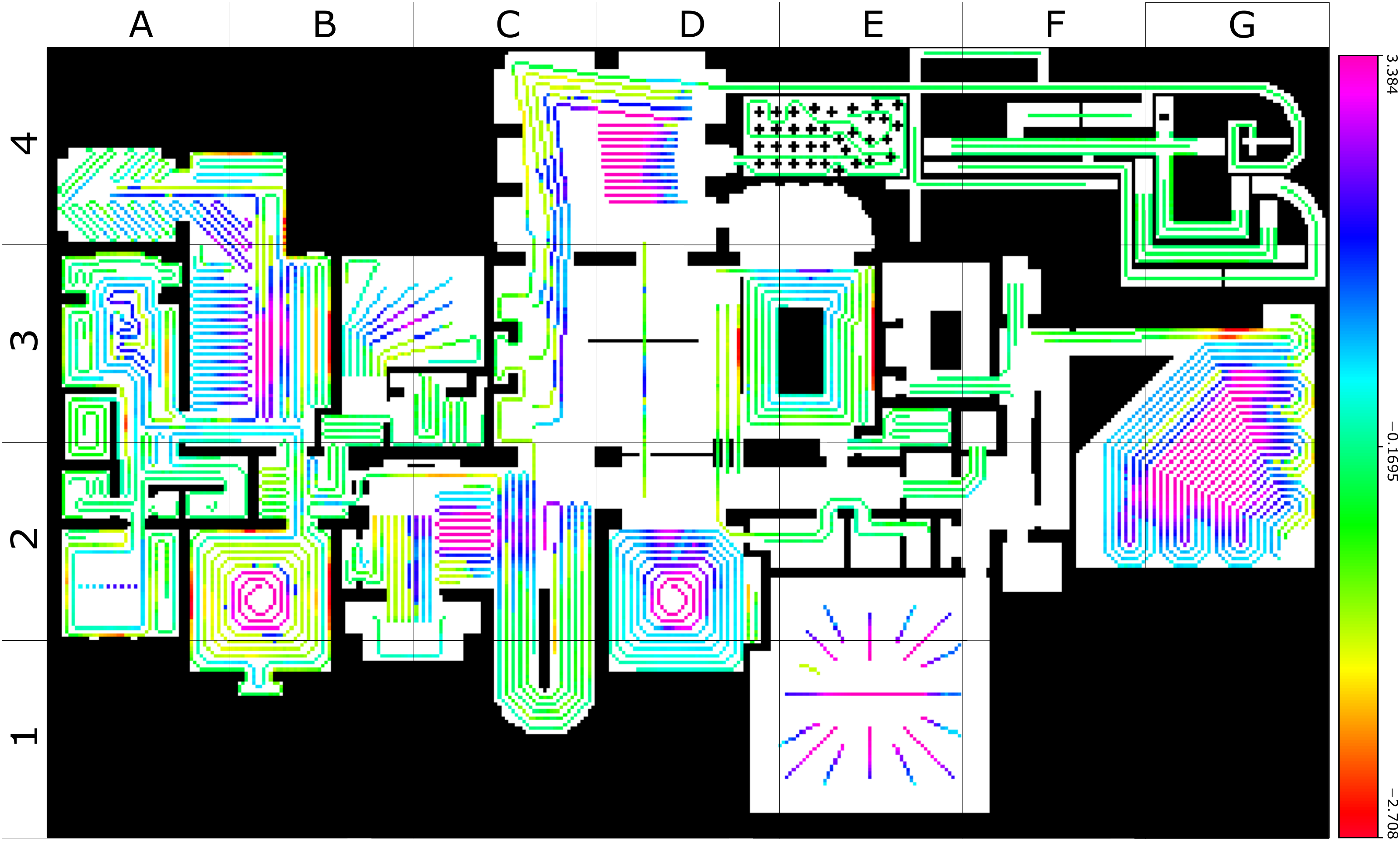}
		\label{fig:VAE_MAP_1_T9}
	}
	\\ 
	\subfloat[
	A Variational Auto-Encoder was trained on isovist-sequences with parameter setting of: $TimeSteps t=9$ \& $Steplength s=2$. The resulting latent space was then sampled 25 times on equally spaced positions between a fixed range. The legend for reconstructed isovist-sequences can be found in Figure \ref{fig:classcolors}.
	]{
		\includegraphics[width=.67\textwidth]{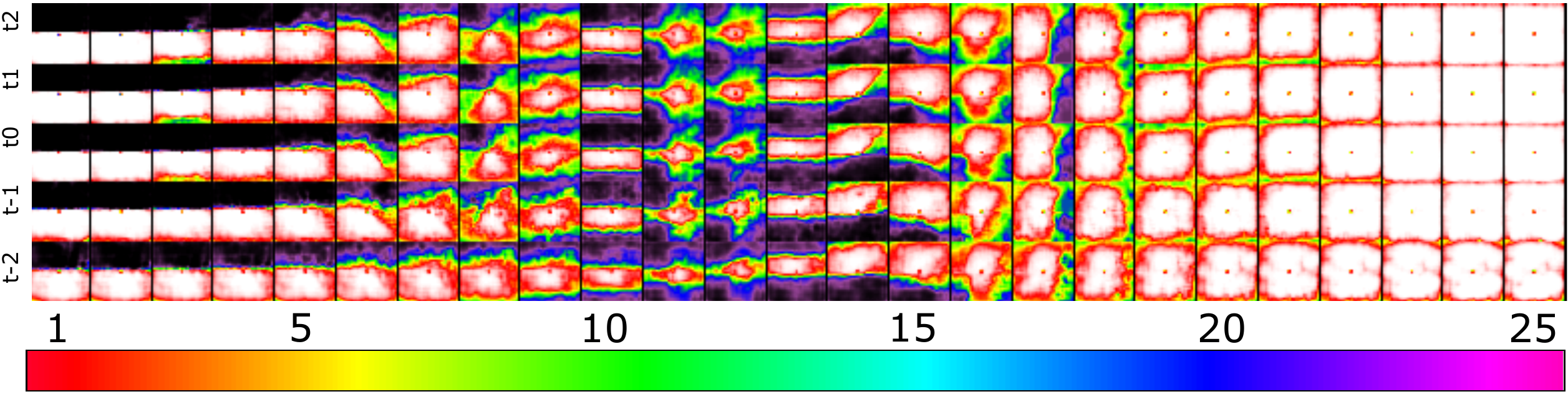}
		\label{fig:VAE_1_T9_EachLatent}
	}
	\caption{Variation of sequence length.}
	\label{fig:seq_len}
\end{figure*}

Our contribution is the design and application of a system that clusters movement along trajectories by spatial perception using isovist in combination with machine learning techniques. 
The training data was built upon isovists as introduced by \cite{benedikt1979take}, which are a reliable and computational cheap tool to represent and measure human spatial perception in the interiors. 
Through several visualizations we have showed the immense potential that lays within such unsupervised trained systems. 
Robotic or assistance systems may be supported to understand movement through spatial structures in foreign environments. 
Future autonomous mobile robots or SLAM systems can produce similar sensory range data, so that our approach may not only be used to annotate trajectories or to help the visual impaired by giving them spatial context, but also to improve existing human-machine-interfaces that rely on discrete situation labels like spoken commands.
In future, our concept could be applied in a three-dimensional domain to measure and process a spatial representation which is closer to the real human world.

Our implementation of the variational auto-encoder was found to generalize along the visited training data. Rather than delivering meaningful human readable classes, the networks seemed to prefer frequently found samples. 
To match the network's prediction with the human spatial perception and semantic labels, a semi-supervised network seems to be the most promising approach for future application. 
For that, the VAE decoder could be extended by a softmax classifier and trained on some labeled data.
Since this work's focus was on showing the feasibility of using the respective ML techniques to learn movement in relation to spatial structures, we did not perform or include a fine tuning of the setup. Future work could cover such an empirical performance study and general concept tuning including e.g. switching from RNN to attention based networks.

Computational costs that come with the training of neural networks increase with higher resolutions and the overall network depth. 
As a consequence, separate training and application units with varying processing power are imaginable.

A small IOT driven robotic unit could be used for online clustering while sending environmental samples to a large-scale GPU or TPU based training unit. 
Such a system would be cheap in application while it keeps learning at the same time. Additionally, the backbone system would benefit from a wide fleet of such cheap robotic units that, all together, draw a vast number of environmental samples. 
A backbone, once trained, could enable large-scale predictions on laptops or even SOC systems. Another solution presented by \cite{lacey2016deep} would be an embedded system acceleration by so called \textit{field programmable gate arrays} (\textit{FPGA}). 
The difference to traditional CPU, GPU, or TPU-based system is that FPGA architectures are tailored for the application in low powered systems \cite{lacey2016deep}.

FPGAs can be thought of as a NN structure built directly into a chip rather than implemented through algorithms that are applied on multi-purpose processing units. Such application-specific hardware acceleration modules have been embedded in large-scale consumer products just recently (e.g., \textit{Google Pixel 2}\footnote{Google Pixel 2 - https://store.google.com/product/pixel\_2\_specs}, \textit{Apple iPhone X}\footnote{Apple iPhone X - https://www.apple.com/iphone-x/specs/}, or \textit{Huawei Mate 10 Pro}\footnote{Mate 10 Pro - http://consumer.huawei.com/en/phones/mate10-pro/}.

Those recent developments move the application of machine learning enhanced processes in our every-daily life.